\pdfoutput=1
\documentclass[11pt]{article}

\usepackage{EMNLP2023}

\usepackage{times}
\usepackage{latexsym}

\usepackage[T1]{fontenc}
\usepackage[utf8]{inputenc}

\usepackage{microtype}

\usepackage{inconsolata}

\usepackage{booktabs}       % professional-quality tables
\usepackage{amsfonts}       % blackboard math symbols
\usepackage{nicefrac}       % compact symbols for 1/2, etc.
\usepackage{microtype}      % microtypography
\usepackage{times}
\usepackage{url}
\usepackage{latexsym}

\usepackage{lipsum}
\usepackage{graphicx}
\usepackage{multirow}
\usepackage{todonotes}
\usepackage{amsmath}
\title{GATology for Linguistics: What Syntactic Dependencies It Knows}

\author{Yuqian Dai, Serge Sharoff, Marc de Kamps \\
        University of Leeds, LS2 9JT, United Kingdom \\ \texttt{\{mlyd,s.sharoff,m.dekamps\}@leeds.ac.uk}}

\begin{document}
\maketitle
\begin{abstract}
Graph Attention Network (GAT) is a graph neural network which is one of the strategies for modeling and representing explicit syntactic knowledge and can work with pre-trained models, such as BERT, in downstream tasks. Currently, there is still a lack of investigation into how GAT learns syntactic knowledge from the perspective of model structure. As one of the strategies for modeling explicit syntactic knowledge, GAT and BERT have never been applied and discussed in Machine Translation (MT) scenarios. We design a dependency relation prediction task to study how GAT learns syntactic knowledge of three languages as a function of the number of attention heads and layers. We also use a paired t-test and F1-score to clarify the differences in syntactic dependency prediction between GAT and BERT fine-tuned by the MT task (MT-B). The experiments show that better performance can be achieved by appropriately increasing the number of attention heads with two GAT layers. With more than two layers, learning suffers. Moreover, GAT is more competitive in training speed and syntactic dependency prediction than MT-B, which may reveal a better incorporation of modeling explicit syntactic knowledge and the possibility of combining GAT and BERT in the MT tasks.
\end{abstract}

\section{Introduction}
The attention mechanism used by many state-of-the-art models can effectively capture potential links between words, as demonstrated by the Transformer model \cite{vaswani2017attention} in different downstream tasks. Inspired by the attention mechanism, \cite{velivckovic2017graph} propose the Graph Attention Network (GAT). In GAT, the update of node features is related to their neighbors, not the whole global state of the network. The attention mechanism also enables it to learn the dependencies between each node and its neighbors adaptively on the graph, which can be applied in transductive and inductive learning. 

One common approach to sentence structure analysis in natural language processing is called syntactic dependency, which uses a tree-like structure to capture dependencies between words in a sentence. Broadly, there are two approaches to modeling such explicit syntactic knowledge. One is represented by RNN variant models such as LSTM or GRU \cite{zhang2019novel,hao2019towards}. However, when dealing with complicated grammatical structures, the dependencies between more distant sentence parts may be beyond the processing range of RNN models. Moreover, some syntactic information may be missed due to information forgetting. The other is based on the attention module in the Transformer model to guide self-attention to specific words \cite{zhang2020sg,mcdonald-chiang-2021-syntax}. All input tokens are still considered when the self-attention mechanism is performed, but strong dependencies between tokens are not explicitly modeled. Also, syntactic knowledge is represented implicitly in the Transformer model and may clash with other modeling requirements, where the model can become a bottleneck.

Unlike RNN and Transformer models, where syntactic knowledge is defined by sequential input, the topological character of GAT simplifies and preserves the structure of syntactic dependencies allowing independent linear information and linguistic knowledge in sentences to be linked via graphs and applied to various downstream tasks. So far, most work has only used GAT to implement the modeling and representation of linguistic knowledge \cite{huang2020syntax,li2022automatic}. Work has yet to discuss how GAT learns syntactic knowledge and whether the number of layers and attention heads influences its syntactic performance, although critical linguistic knowledge represented via GAT is beneficial. And while GAT and the pre-trained model BERT \cite{devlin-etal-2019-bert} are widely used in downstream tasks, there is still a lack of discussion on how GAT and BERT represent syntactic knowledge in Machine Translation (MT) tasks. What are the syntactic knowledge advantages of GAT over BERT fine-tuned for MT tasks? Can an explicit syntactic incorporation strategy based on GAT be used in the MT scenario with BERT? Improving the interpretability of GAT in terms of syntactic knowledge helps to better understand the possibilities of combining graph neural networks and pre-trained language models in MT tasks, including but not limited to BERT. In this work, we investigate the predictions of GAT on syntactic knowledge. We select dependency relations from three languages as our prediction targets in a dependency prediction task to explore whether the number of attention heads and layers in GAT constrains syntactic dependencies. In addition, we also add and design another dependency relation prediction task for BERT fine-tuned for the MT task. Paired t-tests and F1-score compare the prediction differences between GAT and BERT for dependency relations to analyze their syntactic features and the potential of explicit syntactic incorporation strategies via GAT in the MT task. Our main contributions are as follows:
\begin{itemize}
\item We explore which configurations of attention heads and model layers perform best for GAT in learning dependency relations for three different languages. Increasing the number of attention heads can help GAT to be optimal in dependency relation prediction. The prediction results are optimal for two layers, contrary to the intuition that the deeper the network, the better the performance. The deeper layers also make it gradually lose the learning of syntactic knowledge, although some dependency relations are unaffected by this.
\item We evaluate the predictions of GAT and the pre-trained model BERT for typical syntactic dependencies and explore the possibility that syntactic differences exist between them, leading to syntactic knowledge cooperation in the MT task. Paired t-tests reveal significant variability in the F1-score of dependency relation prediction between GAT and BERT fine-tuned by the MT task (MT-B). Although GAT does not have as complex a model structure as BERT, it is competitive in terms of training speed and prediction of syntactic dependencies compared with MT-B in all three different languages. However, GAT fails to predict some dependency relations for each language, and the sample size can constrain its detection.
\end{itemize}

\section{Related Work}
Linguistic knowledge can often be modeled and represented on graphs in natural language processing tasks, e.g., semantic and syntactic information. GAT is a graph neural network that uses an attention mechanism to create a graph across a spatial domain. This mechanism aggregates data from surrounding nodes and determines the relative importance of neighbors to provide new features for each node. It has attracted much interest since it can be used with inductive and transductive learning \cite{salehi2019graph,busbridge2019relational}. So far, most work has focused only on applying syntactic knowledge by GAT in downstream tasks. It is unclear how it represents syntactic knowledge and how model structures, e.g., model layers and attention heads, contribute to syntactic knowledge learning. 

Also, given that GAT can represent explicit linguistic knowledge in different downstream tasks, its integration with the pre-trained model BERT has attracted the most research focus. \cite{huang2020syntax} inject syntactic cognitive knowledge into the model using GAT representation of syntactic knowledge and BERT pre-trained knowledge, which results in better interaction between context and aspectual words. While employing BERT to obtain representations of emotions and contexts, \cite{Li2021SpanLevelEC} use GAT to gather structural data about contexts in the span-level emotion cause analysis task. \cite{Ma2020EntityAwareDD} use graph features and word embeddings to model and represent linguistic knowledge to classify the comparative preference between two given entities. \cite{brody2021attentive} proposes new dynamic attention in GAT but lacks tests of linguistic knowledge. How GAT and BERT interact regarding syntactic knowledge is still being determined, although combining them into downstream tasks can improve performance. Most of the studies have concentrated on discussing and exploring linguistic knowledge in BERT \cite{clark2019does,papadimitriou2021deep}, while the representation of such knowledge in GAT remains unclear. Although some works try to use syntactic knowledge for MT tasks \cite{Peng2021BoostingNM,McDonald2021SyntaxBasedAM}, they do not discuss the possibilities of GAT. \cite{dai-kamps-sharoff:2022:LREC} points out that BERT acts as an MT engine for the encoder to produce low-quality translations when translating sentences with partial syntactic structures, although BERT has syntactic knowledge. Explicit syntactic knowledge benefits MT engines. However, syntactic trees are mostly represented linearly, leading to translation models with missing structural information and information discrimination. Suppose a lightweight GAT can efficiently represent syntactic information topologically and serve as a new strategy to incorporate explicit syntactic knowledge. Its fusion with BERT might improve translation performance and bring more interpretability regarding language knowledge and pre-trained models.

\section{Methodology}
\subsection{Syntactic Learning through Attention Heads and Layers}
\label{GAT}
We use GAT \cite{velivckovic2017graph} as our experimental model to explore how attention heads and layers affect its learning of syntactic knowledge. The node features given to a GAT layer are $X=[x_{1},x_{2},\dotso x_{i},x_{i+1}], x_{i}\in\mathbb{R}^{F}$, where $x$ is the node representing each token in the sentence, ${F}$ is the hidden state of each node given. The Equation (1) and (2) summarise the working mechanism of GAT.
\begin{equation}
\small
h_{i}^{out}=\mathop{\parallel} \limits_{k=1}^K\sigma\left(\displaystyle\sum\limits_{j\in  N_{i}}\alpha_{ij}^{k}W^{k}x_{j}\right)
\end{equation}
\begin{equation}
\small
\alpha_{ij}^{k} = \frac{exp(LeakyReLU(a^{T}[Wx_{i} \parallel Wx_{j}]))}{\sum_{v\in N_{i}}exp(LeakyReLU(a^{T}[Wx_{i} \parallel Wx_{v}]))}
\end{equation}
1-hop neighbors $j\in N_{i}$ for node ${i}$, $\mathop{\parallel} \limits_{k=1}^K$ means the ${K}$ multi-head attention outputs are concatenated in this term, $\sigma$ is a sigmoid function, $h_{i}^{out}$ is the output hidden state of the node ${i}$. $\alpha_{ij}^{k}$ is an attention coefficient between node ${i}$ and ${j}$ with the attention head ${k}$, $W^{k}$ is linear transformation matrix, $a$ is the context vector during training, and $LeakyReLU$ is as activation function \cite{maas2013rectifier}. For simplicity, the feature propagation in GAT can be written as $H_{l+1}=GAT(H_{l},A;\Theta_{l})$, where $H_{l+1}$ is the stacked hidden states of all input nodes at layer ${l+1}$, $A\in\mathbb{R}^{n \times n}$ is the graph adjacency matrix in GAT. $\Theta_{l}$ are the model parameters at that layer.

Each word in a sentence is treated as a graph node, and the edges between the nodes are syntactic dependencies obtained from the Parallel Universal Dependencies (PUD) corpus. GAT needs to predict the dependency relations based on the information of nodes and edges. While syntactic dependencies in linguistics are unidirectional, from parent to child nodes, we treat syntactic dependencies as bidirectional graphs in GAT, from parent to child and from child to parent nodes, respectively. This is because nodes with connectivity have different meanings when they are parent or child nodes, and GAT needs to learn such information to better determine the dependency relations between nodes.

We do not rely on any parser to construct and receive syntactic information of sentences since PUD is a corpus with gold linguistic knowledge, such as lexical information, syntactic dependencies, and other morphological knowledge. In order to reduce the issues with single-language trials, we choose Chinese (Zh), German (De), and Russian (Ru) as the experimental languages and their dependency relations for the tests. The PUD corpus for each language (Chinese PUD\footnote{\url{https://github.com/UniversalDependencies/UD_Chinese-PUD}}, Russian PUD\footnote{\url{https://github.com/UniversalDependencies/UD_Russian-PUD}}, German PUD\footnote{\url{https://github.com/UniversalDependencies/UD_German-PUD}}) has 1,000 sentences (sentences with the same semantics but different languages) that are always arranged in the same order. Constrained by syntactic dependencies, sentences do not follow a sequence on the graph, but a syntactic tree topology provides basic graph structure information.

We increase the number of attention heads and model layers of GAT and assess how well it performs in predicting the dependency relations of different languages under different collocations. We utilize the F1-score as an evaluation metric to indicate how well GAT predicts dependency relations. The number of attention heads of GAT is set to 2, 4, 6, and 8 during experiments, and the number of layers is set to 2, 3, 4, 5, and 6. We record the F1-score of GAT predictions of dependency relations when these parameters are paired with each other. Each language has a training set, validation set, and test set that are each randomly divided into 800, 100, and 100 sentences, respectively. The learning rate = 2e-5, the dropout = 0.2, Adam is the optimizer, and word embeddings = 768.

\subsection{Syntactic Difference with Fine-tuned BERT}
%BERT is often used as a popular pre-trained model for downstream tasks in natural language processing and has achieved significant performance breakthroughs \cite{reimers2019sentence,zhang2019bertscore}.
The fusion of GAT and BERT, attention mechanisms as feature extraction for each model, is possible in downstream tasks, where GAT typically works as an explicit syntactic knowledge incorporation strategy. Given the feasibility of explicit syntactic knowledge represented by GAT,  the possibility exists for explicit knowledge from GAT and implicit knowledge from BERT to improve translation quality. However, there is still a lack of investigation on whether GAT can help and work with BERT in MT scenarios regarding syntactic knowledge. Therefore, we investigate their prediction differences, as well as the interpretability and cooperation potential regarding syntactic knowledge in MT tasks using dependency relation prediction tasks.

Following \cite{dai-kamps-sharoff:2022:LREC}, since we are not limited to one MT task scenario, we choose Chinese (Zh), Russian (Ru), and German (De) as source languages and English (En) as the target language. We use the corresponding BERT-base versions for each source language as an encoder in the MT engine \cite{kuratov2019adaptation,cui2021pre,devlin-etal-2019-bert}. We initially fine-tune BERT for the PUD corpus via a following designed dependency relation prediction task and then for the MT task (MT-B) to ensure that BERT learns the linguistic knowledge from the MT task. Although the pre-training strategies of BERTs are different for each language, their model structures are the same (12 layers and 12 attention heads). The Zh→En and Ru→En MT engines are trained by the United Nations Parallel Corpus (UNPC)\footnote{\url{https://opus.nlpl.eu/UNPC.php}} \cite{ziemski-etal-2016-united}, whereas the De→En MT engine is trained by Europarl\footnote{\url{https://opus.nlpl.eu/Europarl.php}} \cite{koehn-2005-europarl}. In each MT engine, BERT is the encoder, and the decoder comes from the vanilla transformer model, where the training set size is 1.2M sentence pairs, and the validation and test sets are 6K.

The BERT is extracted separately after the fine-tuning of the MT task, and that dependency relation prediction task is applied for BERT again based on the PUD corpus. Inspired by \cite{papadimitriou21deep}, a simple fully-connected layer is added to the last layer of the fine-tuned BERT. Except for the last fully-connected layer, all parameters of BERT are frozen to prevent learning new syntactic knowledge from the PUD corpus. BERT needs to predict the dependency relation corresponding to each token in the sentence. However, BERT and GAT are different in the way they predict dependency relations. GAT is a topology-based prediction and learns explicit syntactic knowledge, therefore, the parent and child nodes in syntactic dependencies are specified. But the dependency relations prediction task for BERT does not provide child nodes but the current parent nodes (the input tokens). Since it is a sequential model that takes into account information from all tokens, this approach simulates as much as possible how it considers syntactic knowledge in the MT tasks. Also, BERT knows the syntactic knowledge since pre-training \cite{htut2019attention,manning2020emergent}. If setting up a complex prediction task, we cannot know whether the knowledge comes from BERT or a complex detection model. Unlike GAT, which always focuses on syntactic knowledge, the syntax is only a part of what BERT needs to learn in the MT tasks. The dependency relation prediction task reveals how BERT knows the syntactic knowledge in the MT scenarios.

We also introduce another BERT model for each language, which only updates the parameters in the dependency relation prediction task for the PUD corpus (UD-B) as a reference model. UD-B is specifically fine-tuned for the PUD corpus, which is considered the best performance of BERT for learning syntactic knowledge. GAT is competitive and has the potential for syntactic knowledge learning if it can beat UD-B on some relations predictions. We evaluate the differences between GAT and BERT in terms of prediction performance in overall and individual terms. First, we use paired t-tests to compare whether there are significant overall differences between GAT and MT-B in their predictions of dependency relations. Second, we discuss the prediction performance of the three models (GAT, MT-B, and UD-B) on individual relations by F1-score to investigate their learning differences in dependency relations.

The dependency relation prediction task of GAT is the same as that of Chapter \ref{GAT}, where GAT has 2 layers and 6 attention heads for Zh, while Ru and De have 4 attention heads. The PUD corpus is the data set of BERTs and GAT. We added K-fold cross-validation to ensure the consistency of the model on the prediction task, where the number of training and test sets are 850 and 150. The F1-score is used as the evaluation metric for the experiments, and the word embeddings = 768, K-fold = 5, learning rate for GAT and BERT = 2e-5, learning rate for fully-connected layer = 1e-4, optimizer = Adam.

\section{Results}
\subsection{Syntactic Predictions with Attention and Layers}
\renewcommand{\thefootnote}{\fnsymbol{footnote}}
As shown in Table \ref{OverallResults}, GAT prefers at least 4 attention heads to obtain the optimal overall prediction performance. The best performance for Ru and De is reached with 2 layers and 4 attention heads, 6 or 8 attention heads yield better prediction outcomes with 2 layers in Zh. In the detailed individual prediction results (see Appendix Sec \ref{GAT test}), the increase of the number of attention heads does help GAT to learn some dependencies. e.g., "\textbf{\textit{cop}}" for Zh, "\textbf{\textit{acl}}" for Ru, and "\textbf{\textit{conj}}" for De. However, the continued adding of attention heads may not lead to more significant performance gains, e.g., when the attention head is over 4 for Ru and De with 2 layers, the further increase does not result in a significant performance gain but a decrease. 

In models like Transformer and BERT, it has been demonstrated that increasing the number of attention heads can improve the model capacity to extract and represent features. This, we believe, is related to the model structure. When sequential input models such as Transformer are utilized, each word in the sentence can contribute to contextual features, improving attention heads can gather and learn probable relationships between words in multiple sub-spaces, resulting in enhanced representations. In contrast to them, where attention mechanism must be allocated to discuss the potential contributions of each token, the perceived range of each word in the sentence is already limited and instructional in GAT due to the structure of syntactic dependency. Thus, the effect of the increase in the number of attention heads is much less pronounced than the gains of the Transformer model. Adding attention heads may also cause redundancy of information, thereby reducing its learning of syntactic knowledge.

\begin{table}[!ht]
\centering
\small
\begin{tabular}{ccccc}
\hline
\multicolumn{1}{l}{} & \multicolumn{4}{c}{\textbf{Zh}}                                                                           \\
\multicolumn{1}{l}{} & \textbf{2 Heads}              & \textbf{4 Heads}              & \textbf{6 Heads}              & \textbf{8 Heads}              \\ 
\textbf{2 Layers}          & 0.63                     & 0.62                     & \textbf{0.64}                     & \textbf{0.64}                    \\
\textbf{3 Layers}          & 0.64                     & 0.61                     & 0.62                     & 0.63                     \\
\textbf{4 Layers}          & 0.56                     & 0.58                     & 0.64                     & 0.49                    \\
\textbf{5 Layers}          & 0.49                     & 0.50                     & 0.51                     & 0.50                    \\
\textbf{6 Layers}          & 0.37                     & 0.40                     & 0.33                     & 0.33                     \\ \hline
\multicolumn{1}{l}{} & \multicolumn{4}{c}{\textbf{Ru}}                                                                           \\
\multicolumn{1}{l}{} & \textbf{2 Heads}              & \textbf{4 Heads}              & \textbf{6 Heads}              & \textbf{8 Heads}              \\ 
\textbf{2 Layers}          & 0.58                     & \textbf{0.61}                     & 0.47                     & 0.56                     \\
\textbf{3 Layers}          & 0.45                     & 0.55                     & 0.54                     & 0.53                     \\
\textbf{4 Layers}          & 0.44                     & 0.47                     & 0.56                     & 0.57                     \\
\textbf{5 Layers}          & 0.42                     & 0.52                     & 0.46                     & 0.49                     \\
\textbf{6 Layers}          & 0.41                     & 0.36                     & 0.31                     & 0.33                    \\ \hline
\multicolumn{1}{l}{} & \multicolumn{4}{c}{\textbf{De}}                       \\
\multicolumn{1}{l}{} & \textbf{2 Heads} & \textbf{4 Heads} & \textbf{6 Heads} & \textbf{8 Heads} \\ 
\textbf{2 Layers}          & 0.64                     & \textbf{0.67}                     & 0.64                     & 0.56                     \\
\textbf{3 Layers}          & 0.60                     & 0.56                     & 0.56                     & 0.57                     \\
\textbf{4 Layers}          & 0.56                     & 0.50                     & 0.53                     & 0.53                     \\
\textbf{5 Layers}          & 0.58                     & 0.61                     & 0.50                     & 0.47                     \\
\textbf{6 Layers}          & 0.48                     & 0.49                     & 0.48                     & 0.42          \\ \hline
\end{tabular}
\caption{Overall GAT predictions of syntactic relationships for three languages with different numbers of attention heads and layers. The increased number of attention heads and layers does not guarantee a performance gain.}
\label{OverallResults} 
\end{table}

\begin{figure}[!ht]
\centering
\includegraphics[scale=0.6]{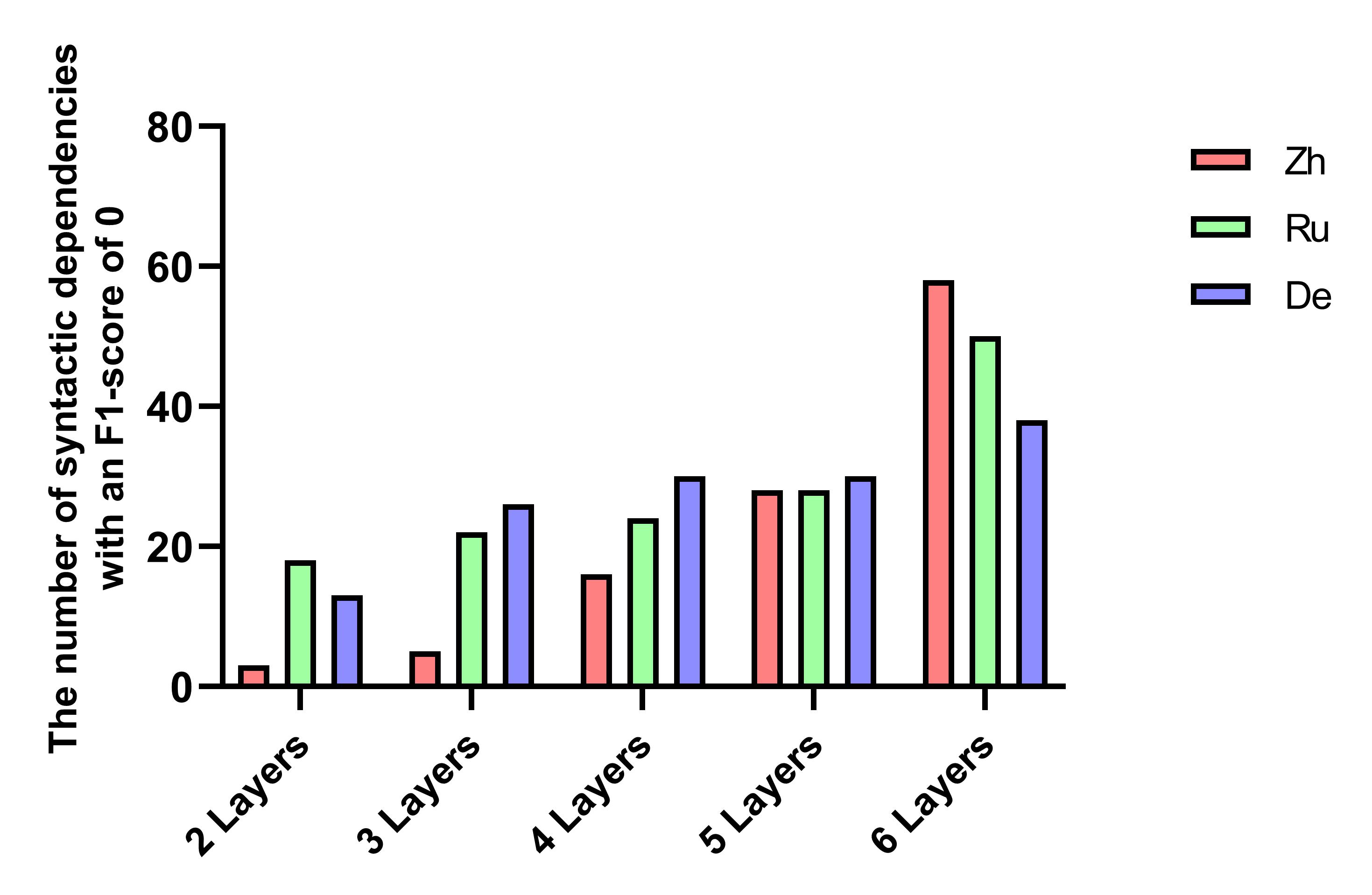}
\caption{The number of F1-score dropped to 0 made by the GAT in different layers with a different number of attention heads. Although each layer has 2, 4, 6, and 8 attention heads, increasing the number of layers invariably results in more failures for syntactic knowledge learning.}
\label{SytaxPrediction} 
\end{figure}

We note that the GAT prediction scores for dependency relations are optimistic with the proper number of attention heads and layers. However, experiments demonstrate that increasing the number of GAT layers significantly reduces overall prediction results (more details are in Appendix Sec \ref{GAT test}), and GAT gradually loses learning and prediction of some dependency relations, as shown in Table \ref{SytaxDominance}. As the number of layers increases, predicting some dependency relations is difficult for GAT, and the F1-score decreases and even drops to 0. We record the number of dependency relations with an F1-score of 0 under the different number of attention heads in each layer for each language, as shown in Figure \ref{SytaxPrediction}. When the number of GAT layers is more than 3, the F1-score of 0 becomes more frequent, and adding attention heads does not solve this problem. The increase of GAT layers does not result in increased performance, which could be because the nodes lose their attributes or absorb some unnecessary information, resulting in a model performance decrease. However, GAT still shows strong prediction performance for some dependency relations, e.g., "\textbf{\textit{flat}}", "\textbf{\textit{compound}}", "\textbf{\textit{nmod}}" in Zh. "\textbf{\textit{cop}}", "\textbf{\textit{flat:name}}", "\textbf{\textit{nummod}}" in Ru, "\textbf{\textit{nmod}}", "\textbf{\textit{obl}}" and "\textbf{\textit{det}}" in De. Such dependency relations do not appear to be 0 for F1-score as the number of layers increases, and they maintain valid prediction scores when the depth of the model reaches 6 layers. Although GAT learns differently for each language, several common dependency relations share a feature that the F1-score never becomes 0: "\textit{\textbf{advmod}}", "\textit{\textbf{case}}", "\textit{\textbf{cc}}", "\textit{\textbf{mark}}", "\textit{\textbf{nsubj}}", "\textit{\textbf{punct}}". It implies that GAT exhibits robust learning of syntactic dependencies either for 2 layers or 6 layers, which explains why explicit syntactic knowledge incorporation strategies via GAT are feasible in downstream tasks. Deeper GAT still learns partial dependency relations, even the same relations in different languages, which may suggest that deeper graph neural networks are possible.

\begin{table*}[!ht]
\centering
\small
\begin{tabular}{cc|ccc|ccc|ccc} 
\hline
\multicolumn{2}{c|}{GAT}   & \multicolumn{3}{c|}{Zh} & \multicolumn{3}{c|}{Ru} & \multicolumn{3}{c}{De}    \\
Layers             & Heads & advmod & clf  & dep     & case & flat & mark      & acl:relcl & cc   & naubj  \\ 
\hline
\multirow{4}{*}{2} & 2     & 0.90    & 0.87 & 0.64    & 0.99 & 0.85 & 0.97      & 0.71      & 0.97 & 0.75   \\
                   & 4     & 0.90    & 0.82 & 0.63    & 0.99 & 0.86 & 0.94      & 0.75      & 0.99 & 0.72   \\
                   & 6     & 0.91   & 0.89 & 0.66    & 0.98 & 0.87 & 0.96      & 0.75      & 0.96 & 0.72   \\
                   & 8     & 0.90    & 0.83 & 0.62    & 0.98 & 0.86 & 0.90       & 0.41      & 0.97 & 0.69   \\ 
\hline
\multirow{4}{*}{3} & 2     & 0.90    & 0.88 & 0.64    & 0.98 & 0    & 0.93      & 0.60       & 0.96 & 0.78   \\
                   & 4     & 0.91   & 0.86 & 0.64    & 0.98 & 0.86 & 0.94      & 0.45      & 0.96 & 0.71   \\
                   & 6     & 0.90    & 0.88 & 0.66    & 0.98 & 0.77 & 0.93      & 0.41      & 0.96 & 0.72   \\
                   & 8     & 0.91   & 0.9  & 0.66    & 0.99 & 0.86 & 0.93      & 0.46      & 0.96 & 0.74   \\ 
\hline
\multirow{4}{*}{4} & 2     & 0.89   & 0.68 & 0.64    & 0.97 & 0    & 0.94      & 0.52      & 0.84 & 0.74   \\
                   & 4     & 0.90    & 0.66 & 0.65    & 0.99 & 0.77 & 0.94      & 0.45      & 0.85 & 0.73   \\
                   & 6     & 0.91   & 0.69 & 0.68    & 0.99 & 0.67 & 0.97      & 0.40       & 0.85 & 0.77   \\
                   & 8     & 0.90    & 0    & 0.64    & 0.99 & 0.8  & 0.94      & 0.45      & 0.96 & 0.74   \\ 
\hline
\multirow{4}{*}{5} & 2     & 0.90    & 0    & 0       & 0.97 & 0.55 & 0.93      & 0.42      & 0.85 & 0.78   \\
                   & 4     & 0.90    & 0    & 0       & 0.98 & 0.77 & 0.96      & 0.68      & 0.82 & 0.79   \\
                   & 6     & 0.90    & 0    & 0       & 0.97 & 0.67 & 0.93      & 0.44      & 0.81 & 0.72   \\
                   & 8     & 0.89   & 0    & 0       & 0.99 & 0.48 & 0.96      & 0.43      & 0.86 & 0.73   \\ 
\hline
\multirow{4}{*}{6} & 2     & 0.83   & 0    & 0       & 0.94 & 0    & 0.91      & 0         & 0.83 & 0.65   \\
                   & 4     & 0.86   & 0    & 0       & 0.95 & 0    & 0.97      & 0         & 0.78 & 0.65   \\
                   & 6     & 0.84   & 0    & 0       & 0.94 & 0    & 0.93      & 0         & 0.79 & 0.67   \\
                   & 8     & 0.86   & 0    & 0       & 0.96 & 0    & 0.93      & 0.37      & 0.85 & 0.63   \\
\hline
\end{tabular}
\caption{The predictions of some syntactic dependencies in three different languages are shown. As the number of layers increases, GAT gradually loses the learning of syntactic dependencies, and even F1-score drops to 0. Some dependencies are unaffected and continue to have relatively high prediction scores.}
\label{SytaxDominance} 
\end{table*}

\subsection{Syntactic Differences with BERT}
As shown in Table \ref{T-TESTResult}, paired t-tests indicate that the p-value is less than the significance level (0.05) in the Zh prediction task (some dependency relations with an F1-score of 0 are considered outliers not in the statistics), which means that the null hypothesis ($H_0$) that there is no difference between GAT and MT-B in the F1-score of dependency relation prediction is rejected. Instead, the alternative hypothesis ($H_1$) that the F1-score of dependency relation prediction between GAT and MT-B is statistically significant is accepted. A similar circumstance occurs when paired t-tests are performed in Ru and De.

\begin{table*}[!ht]
\centering
\small
\begin{tabular}{c|c|c|c|c|c|c|c} 
\hline
\multicolumn{1}{l|}{\textbf{Languages}} & \multicolumn{1}{l|}{\textbf{Observations}} & \textbf{Sample size} & \textbf{Significance level} & \textbf{Mean} & \textbf{STDev} & \textbf{T-value}       & \textbf{P-value}        \\ 
\hline
\multirow{2}{*}{Zh}                     & MT-B                                                                  & \multirow{2}{*}{31}  & \multirow{6}{*}{0.05}       & 0.6           & 0.2            & \multirow{2}{*}{3.450}  & \multirow{2}{*}{0.001}  \\
                                        & GAT                                                                      &                      &                             & 0.7           & 0.3            &                        &                         \\ 
\cline{1-3}\cline{5-8}
\multirow{2}{*}{Ru}                     & MT-B                                                                  & \multirow{2}{*}{28}  &                             & 0.7           & 0.2            & \multirow{2}{*}{2.283} & \multirow{2}{*}{0.030}  \\
                                        & GAT                                                                      &                      &                             & 0.7           & 0.2            &                        &                         \\ 
\cline{1-3}\cline{5-8}
\multirow{2}{*}{De}                     & MT-B                                                                  & \multirow{2}{*}{27}  &                             & 0.6           & 0.2            & \multirow{2}{*}{2.062} & \multirow{2}{*}{0.049}  \\
                                        & GAT                                                                      &                      &                             & 0.7           & 0.3            &                        &                         \\
\hline
\end{tabular}
\caption{Paired t-tests are used to compare the findings of GAT and MT-B on syntactic dependency prediction. There is a significant difference in the prediction results between the two models.}
\label{T-TESTResult} 
\end{table*}

\begin{table*}[!ht]
\centering
\small
\begin{tabular}{l|r|cc|c|r|cc|c|r|cc|c} 
\hline
              & \multicolumn{4}{c|}{\textbf{Zh}}                            & \multicolumn{4}{c|}{\textbf{Ru}}                            & \multicolumn{4}{c}{\textbf{De}}                              \\
              & \textbf{\#} & MT-B  & \multicolumn{1}{c}{GAT} & UD-B & \textbf{\#} & MT-B  & \multicolumn{1}{c}{GAT} & UD-B & \textbf{\#} & MT-B  & \multicolumn{1}{c}{GAT} & UD-B  \\ 
\cline{2-13}
acl           & 20       & 0     & 0                       & \textit{0}     & 256      & \textbf{0.523} & 0.392                   & \textit{0.854} & 20       & 0     & 0                       & \textit{0}      \\
acl:relcl     & 448      & 0.420 & \textbf{0.913}                   & 0.836 & 160      & \textbf{0.451} & 0.405                   & \textit{0.960}  & 271      & \textbf{0.659} & 0.605                   & \textit{0.912}  \\
advcl         & 516      & 0.279 & \textbf{0.376}                   & \textit{0.728} & 197      & 0.330 & \textbf{0.334}                   & \textit{0.842} & 221      & 0.414  & \textbf{0.495}                   & \textit{0.832}  \\
advmod        & 1225     & 0.668  & \textbf{0.909}                   & \textit{0.946} & 914      & 0.843 & \textbf{0.902}                   & \textit{0.964} & 1120     & 0.622 & \textbf{0.984}                   & 0.958  \\
amod          & 419      & 0.400 & \textbf{0.919}                   & 0.874 & 1791     & 0.872 &\textbf{0.979}                   & \textit{0.982} & 1101     & 0.658 & \textbf{0.935}                   & \textit{0.976}  \\
appos         & 248      & \textbf{0.480} & 0.423                   & \textit{0.740} & 121      & 0.428 & \textbf{0.436}                   & \textit{0.570}  & 265      & 0.350 & \textbf{0.561}                   & \textit{0.786}  \\
aux           & 680      & 0.758 & \textbf{0.875}                   & \textit{0.966} & 42       & \textbf{0.878} & 0.836                   & \textit{0.932} & 367      & 0.818  & \textbf{0.862}                   & \textit{0.972}  \\
aux:pass      & 79       & \textbf{0.862} & 0                       & \textit{0.970} & 128      & 0.958 & \textbf{0.988}                   & 0.968 & 230      & 0.835 & \textbf{0.934}                   & \textit{0.965}  \\
case          & 1319     & 0.734 & \textbf{0.963}                   & 0.928 & 2121     & 0.931 & \textbf{0.983}                   & 0.981 & 2055     & 0.840 & \textbf{0.994}                   & 0.986  \\
case:loc      & 346      & 0.670 & \textbf{0.779}                   & \textit{0.954} & -        & -     & -                       & -     & -        & -     & -                       & -      \\
cc            & 283      & 0.851 & \textbf{0.990}                   & 0.938 & 599      & 0.954  & \textbf{0.969}                   & \textit{0.988} & 723      & 0.829 & \textbf{0.981}                   & 0.972  \\
ccomp         & 403      & 0.148 & \textbf{0.277}                   & \textit{0.656} & 132      & 0.469 & \textbf{0.536}                   & \textit{0.752} & 169      & \textbf{0.289} & 0.296                   & \textit{0.704}  \\
clf           & 357      & \textbf{0.816} & 0.737                   & \textit{0.980}  & -        & -     & -                       & -     & -        & -     & -                       & -      \\
compound      & 1777     & 0.619 & \textbf{0.881}                   & \textit{0.886} & 9        & 0     & 0                       & 0     & 250      & 0.465 & \textbf{0.496}                   & \textit{0.850}  \\
conj          & 383      & 0.481 & \textbf{0.976}                   & 0.842 & 695      & 0.732 & \textbf{0.862}                   & \textit{0.920}  & 841      & 0.591 & \textbf{0.673}                   & \textit{0.912}  \\
cop           & 196      & 0.588 & \textbf{0.962}                   & 0.842 & 87       & 0.756 & \textbf{0.983}                   &0.830  & 275      & \textbf{0.782} & 0.755                   & \textit{0.954}  \\
% csubj         & 72       & 0     & 0                       & 0.156 & 47       & 0     & 0                       & 0.782 & 28       & 0     & 0                       & 0      \\
dep           & 396      & 0.251 & \textbf{0.556}                   & \textit{0.742} & -        & -     & -                       & -     & -        & -     & -                       & -      \\
det           & 338      & 0.712 & \textbf{0.963}                   & 0.956 & 476      & 0.870 & \textbf{0.997}                   & 0.974 & 2760     & 0.914 & \textbf{0.996}                   & 0.980  \\
% discourse:sp & 87        & \textbf{0.814} & 0.173                   & 0.980 & -        & -     & -                       & -     & -        & -     & -                       & -      \\
expl         & -        & -     & -                       & -     & 7        & 0     & 0                       & \textit{0.890}  & 90       & \textbf{0.711}  & 0.319                   & \textit{0.982}  \\
fixed        & -        & -     & -                       & -     & 222      & \textbf{0.600} & 0.577                   & \textit{0.846} & 7        & 0     & 0                       & \textit{0}      \\
flat         & 91       & 0.724 & \textbf{0.867}                   & \textit{0.965} & 61       & 0.220 & \textbf{0.583}                   & \textit{0.538} & 4       & 0.080  & \textbf{0.371}                   & \textit{0.344}  \\
flat:foreign  & -        & -     & -                       & -     & 97       & 0.330  & \textbf{0.903}                   & 0.892 & -        & -     & -                       & -      \\
flat:name    & 142      & 0.791 & \textbf{0.897}                   & \textit{0.936} & 222      & \textbf{0.910}  & 0.888                   & \textit{0.986} & 164      & 0.486 & \textbf{0.844}                   & 0.762  \\
iobj         & 15       & 0     & 0                       & \textit{0.134} & 190      & \textbf{0.510} & 0                       & \textit{0.730}  & 95       & \textbf{0.494}  & 0                       & \textit{0.874}  \\
mark         & 291      & 0.512 &\textbf{ 0.980}                   & 0.905   & 287      & 0.780 & \textbf{0.867}                   & 0.854 & 459      & 0.817 & \textbf{0.992}                   & 0.980  \\
mark:adv     & 22       & \textbf{0.992}  & 0.400                   & \textit{0.970}  & -        & -     & -                       & -     & -        & -     & -                       & -      \\
mark:prt     & 338      & \textbf{0.438} & 0.237                   & \textit{0.838} & -        & -     & -                       & -     & -        & -     & -                       & -      \\
mark:relcl   & 626      & \textbf{0.869} & 0.756                   & \textit{0.944} & -        & -     & -                       & -     & -        & -     & -                       & -      \\
nmod         & 707      & 0.386  & \textbf{0.919}                   & 0.826 & 1934     & 0.667 & \textbf{0.870}                   & \textit{0.920}  & 1099     & 0.590  & \textbf{0.749}                   & \textit{0.888}  \\
nsubj         & 1772     & 0.598 & \textbf{0.612}                   & \textit{0.906} & 1362     & \textbf{0.719} & 0.666                   & \textit{0.936} & 1482     & 0.659 & \textbf{0.678}                   & \textit{0.950}  \\
nsubj:pass   & 71       & \textbf{0.127} & 0                       & \textit{0.766} & 186      & \textbf{0.280} & 0                       & \textit{0.904} & 207      & \textbf{0.391}  & 0                       & \textit{0.974}  \\
nummod        & 809      & 0.848 & \textbf{0.993}                   &0.988 & 183      & 0.529  & \textbf{0.690}                   & \textit{0.732} & 227      & 0.736 & \textbf{0.808}                   & \textit{0.926}  \\
% nummod:gov   & -        & -     & -                       & -     & 73       & \textbf{0.486} & 0                       & 0.572 & -       & -    & -                       & -      \\
obj          & 1526     & 0.459 & \textbf{0.558}                   & \textit{0.858} & 749      & \textbf{0.558}  & 0.518                   & \textit{0.928} & 898      & \textbf{0.599} & 0.485                   & \textit{0.960}  \\
obl          & 686      & 0.204 & \textbf{0.846}                   & 0.738 & 1465     & 0.672 &\textbf{0.911}                   & \textit{0.914} & 1304     & 0.584 & \textbf{0.821}                   & \textit{0.918}  \\
obl:agent     & 22       & \textbf{0.364} & 0                       & \textit{0.888} & 12       & 0     & 0                       & \textit{0.520} & -        & -     & -                       & -      \\
obl:patient   & 39       & 0     & 0                       & \textit{0.986} & -        & -     & -                       & -     & -        & -     & -                       & -      \\
obl:tmod     & 214      & \textbf{0.534} & 0.104                   & \textit{0.816} & -        & -     & -                       & -     & 119       & \textbf{0.623} & 0.216                   &\textit{0.832}  \\
% orphan       & -        & -     & -                       & -     & 17       & 0     & 0                       & 0     & 10       & 0     & 0                       & 0      \\
parataxis    & -        & -     & -                       & -     & 195      & \textbf{0.525} & 0.200                   & \textit{0.706} & 68       & 0.160    & 0                       & \textit{0.524}  \\
punct        & 2902     & 0.754 & \textbf{0.990}                   & 0.990 & 2977     & 0.960 & \textbf{0.990}                   & 0.990 & 2771     & 0.932 & \textbf{0.999}                   &0.981  \\
root         & 1000     & 0.493 & \textbf{0.968}                   & 0.894 & 1000     & 0.886 & \textbf{0.994}                   & 0.982 & 1000     & 0.711 & \textbf{0.932}                   & \textit{0.982}  \\
xcomp         & 537      & 0.292 & \textbf{0.437}                   &\textit{0.804} & 331      & 0.591 & \textbf{0.634}                   & \textit{0.880} & 190      & \textbf{0.430} & 0.291                   & \textit{0.820}  \\
\hline
\end{tabular}
\caption{Prediction scores of GAT, MT-B, and UD-B for dependency relations based on PUD corpus. GAT is more competitive than MT-B in predicting most dependency relations, shown in bold format, and some relations can surpass UD-B, shown in the non-italic format in the column of UD-B.}
\label{GATandBERT} 
\end{table*}

Investigating the prediction of each dependency relation based on the F1-score as shown in Table \ref{GATandBERT}, we find that GAT dominates the prediction of the vast majority of dependency relations with higher F1-score, with only a small proportion losing out to that of MT-B. We argue that although BERT is fine-tuned by the PUD corpus and MT task, its learning of syntactic knowledge is still inadequate in this case. BERT may produce similar results under fine-tuning in other downstream tasks since many studies have shown that incorporating syntactic knowledge through GAT with BERT in downstream tasks can improve performance \cite{huang2020syntax,chen2021combining,zhou2022dynamic}. If BERT would remain highly aware of syntactic knowledge after fine-tuning, then explicit syntactic incorporation strategies via GAT would hardly have a positive substantial impact in downstream tasks.

The study of \cite{dai-kamps-sharoff:2022:LREC} finds that when detection of syntactic dependencies deteriorates, MT quality drops, where the dependency relations can be "\textit{\textbf{appos}}", "\textit{\textbf{case}}", "\textit{\textbf{flat}}", "\textit{\textbf{flat:name}}", and "\textit{\textbf{obl}}". Experiments show that GAT is superior in learning and predicting these dependencies compared to MT-B in three languages, which may support the application of explicit syntactic knowledge incorporation strategy via GAT in MT scenarios. Moreover, GAT dominates MT-B in predicting certain dependencies, e.g., "\textbf{\textit{conj}}", "\textbf{\textit{nmod}}" in Chinese, "\textbf{\textit{cop}}", "\textbf{\textit{obl}}" in Russian, and "\textbf{\textit{advmod}}", "\textbf{\textit{flat:name}}" in German. Also, the relation of "\textit{\textbf{root}}" as the sentence main predicate\footnote[1]{One of the orphaned dependents gets promoted to the root position if the main predicate is absent.} is the root node and is used to express the main substance in a sentence. Since it appears in every sentence, GAT and BERT predict differently and cannot be linked to a drop in MT quality, the fact that GAT is better in detecting compared with MT-B, means that BERT fine-tuned for the PUD corpus and MT task still lack the ability to detect. Also, GAT has better predictive performance in most cases, where GAT is more competitive for 25 of the 37 dependency relations in Zh, 20 out of 33 relations in Ru, and 20 out of 32 relations in De. The main role of GAT in the MT task is to learn and represent the syntactic information provided by the parser. Suppose GAT can represent syntactic knowledge as correctly as possible and provide such knowledge to the translation engine. The translation results may become more fluent and natural, which also provides the possibility of incorporating explicit syntactic knowledge via GAT into the MT task more effectively.

Most dependency relations have less than 500 samples, indicating that the training sample cost of GAT is not expensive compared with BERT pre-trained for a large corpus. The same number of training samples can outperform MT-B in most syntactic dependencies and UD-B in a few cases. But when the number of samples is much smaller (less than 100), learning language knowledge is challenging for both BERT and GAT. Benefiting from pre-training and a more robust model structure, BERT can somewhat alleviate this problem. However, GAT cannot. There are 8 dependency relations with less than 100 in Zh, and the number of undetectable ones in GAT is 6: "\textbf{\textit{acl}}", "\textbf{\textit{aux:pass}}", "\textbf{\textit{iobj}}", "\textbf{\textit{nsubj:pass}}", "\textbf{\textit{obl:agent}}", "\textbf{\textit{obl:patient}}". Ru and De contain 7, respectively, where the number of failed detections is 3 and 4. They are "\textbf{\textit{compound}}", "\textbf{\textit{expl}}", "\textbf{\textit{obl:agent}}" in Ru, and "\textbf{\textit{acl}}", "\textbf{\textit{fixed}}", "\textbf{\textit{iobj}}", "\textbf{\textit{parataxis}}" in De. Besides, specific dependency relations is difficult for GAT. "\textbf{\textit{iobj}}" and "\textbf{\textit{nsubj:pass}}" in the three languages cannot be predicted by GAT. These two relations are consistent in linguistic knowledge classification, with core arguments as functional categories and nominals as structural categories. GAT may lack sufficient learning of the syntactic subjects of indirect objects and passive clauses. However, achieving robust syntactic dependency learning and obtaining acceptable performance for three languages with only several times fewer model parameters than BERT without sacrificing training speed (see Appendix Sec \ref{GAT and BERT parameters}), a lightweight and inexpensive GAT is competitive enough in modeling explicit syntactic knowledge.

UD-B performs best in terms of the F1-score, given that BERT is pre-trained with a large amount of data and is more complicated than GAT regarding the number of attention heads and the model structure, the prediction results are not surprising. But it does not obtain the highest scores for all predictions of dependency relations, GAT still outperforms some, e.g., "\textit{\textbf{conj}}" in Zh, "\textit{\textbf{det}}" in Ru, and "\textit{\textbf{advmod}}" in De. There are a total of 8 dependency relations in Zh where GAT outperforms UD-B, with 6 of them having a sample size higher than 300. There are 7 of them in Ru, 3 of which are over 300. Also, there are 8 in De, 6 of which are over 300. In addition, we record the common relations that outperformed UD-B in prediction in all three languages: "\textit{\textbf{case}}", "\textit{\textbf{mark}}", "\textit{\textbf{det}}", and "\textit{\textbf{cc}}". The better identification of cross-linguistic dependency relations suggests that GAT has better knowledge and mastery of them, even though it is not pre-trained. Such features may allow certain linguistic-specific knowledge to be better applied in MT scenarios through explicit syntactic knowledge incorporation strategies via GAT. Also, the three dependency relations,"\textit{\textbf{case}},"\textit{\textbf{cc}}, and "\textit{\textbf{mark}}, are common to all three languages and are not affected by the increase in the number of layers, which results in an F1-score of 0. It implies that GAT may have developed cross-linguistic knowledge, although only in small parts.

\section{Conclusions}
This study investigates how GAT learns syntactic knowledge and the effect of attention heads and model layers. GAT prefers at least 4 attention heads to learn syntactic knowledge. However, when the number of layers exceeds 2, GAT gradually loses the learning of syntactic dependencies. We also investigate the possibility of fusing GAT and BERT in MT scenarios. Paired t-tests and F1-score indicate statistically significant differences in dependency relation prediction between GAT and MT-B. GAT maintains competitive in modeling and learning of syntactic dependencies without sacrificing training speed compared with MT-B. It even outperforms UD-B in learning a small number of syntactic dependencies. However, GAT fails to detect some dependency relations and suffers from sample size. Future study will include research on the fusion of syntactic knowledge via GAT and BERT to improve the translation quality in MT tasks.

\section{Limitations}
In this work, we find that as the number of layers increases, the F1-score of 0 is obtained for some dependency relations in GAT. However, the lack of explainability of such a phenomenon still leaves gaps in the investigation. Also, the PUD corpus for each language contains 1,000 syntactic annotated sentences. It does not provide a sufficient number for all dependency relations in the experiment, making the experiment have to discard some dependency relations in the prediction.

\bibliography{anthology,custom}
\bibliographystyle{acl_natbib}

\appendix
\clearpage
\section{Appendix}
\label{sec:appendix}
\subsection{Syntactic Predictions with Attention and Layers}
\label{GAT test}
We investigate syntactic dependency learning in GAT for Chinese (Zh), Russian (Ru), and German (De) for different numbers of attention heads (A) and layers (L) as shown in Table \ref{GATandBERT-1} to Table \ref{GATandBERT-5}. As some dependency relations in the PUD corpus are uncommon with only a small number of samples, they do not reasonably reflect the learning performance of the model, we remove them in the experiments. Due to the diversity of linguistic knowledge, the categories of syntactic dependencies may vary between languages.

\subsection{Comparison of the relevant parameters of BERT and GAT}
\label{GAT and BERT parameters}
We record the GAT, MT-B, and UD-B comparisons regarding model parameters and training speed. In our study, we follow \cite{velivckovic2017graph} where batch size = 1. To fairly compare the differences between GAT and BERT, the batch size of BERT is not only 16, but we also set it to 1. As shown in Table \ref{GATandBERT speed}, with both batch sizes of 1, the training speed of the lightweight GAT and MT-B on each epoch is similar but far outperforms UD-B. It reveals that GAT still obtains better learning of syntactic knowledge with fewer model parameters and without sacrificing training speed. Although UD-B obtains the best performance on the prediction of syntactic dependencies, it has the slowest training speed. In the downstream task, fine-tuning BERT is more costly because it focuses on more than just syntactic knowledge.

\begin{table}[!ht]
\centering
\small
\begin{tabular}{l|c|cccc} 
\hline
                      & GAT       & \multicolumn{2}{c}{MT-B} & \multicolumn{2}{c}{UD-B}  \\ 
\hline
Batch size            & 1         & 16  & 1                  & 16  & 1                   \\
Speed (sec per epoch) & 8         & 1.5 & 7.5                & 3.5 & 28                  \\
Parameters for Zh     & 5,439,021 & \multicolumn{4}{c}{102,303,022}                      \\
Parameters for Ru     & 7,345,296 & \multicolumn{4}{c}{177,884,969}                      \\
Parameters for De     & 6,401,324 & \multicolumn{4}{c}{109,115,949}                      \\
\hline
\end{tabular}
\caption{Comparison of GAT, MT-B, and UD-B in terms of model parameters and training speed.}
\label{GATandBERT speed} 
\end{table}

\begin{table*}[!ht]
\centering
\small
\begin{tabular}{c|cccccccccc} 
\hline
\multicolumn{11}{c}{Zh}                                                                                                                                                                                                                                              \\
\multicolumn{1}{c}{L-A} & acl:relcl            & advcl                & advmod               & amod                 & appos                & aux                  & case                 & case:loc             & cc                   & ccomp                 \\ 
\hline
2--2                           & 0.82                 & 0                    & 0.90                  & 0.80                  & 0.60                  & 0.90                  & 0.98                 & 0.95                 & 0.99                   & 0.41                     \\
2--4                           & 0.83                 & 0                    & 0.90                  & 0.81                 & 0.55                 & 0.91                 & 0.99                 & 0.94                 & 0.99                    & 0.40                  \\
2--6                           & 0.87                 & 0.14                 & 0.91                 & 0.85                 & 0.61                 & 0.91                 & 0.99                 & 0.91                 & 0.99                    & 0.53                  \\
2--8                          & 0.84                 & 0.15                 & 0.90                 & 0.80                 & 0.58                 & 0.91                 & 0.99                 & 0.94                 & 0.99                 & 0.30                  \\
3--2                           & 0.87                 & 0                    & 0.90                 & 0.84                 & 0.54                 & 0.90                  & 0.99                 & 0.92                 & 0.99                    & 0.66                  \\
3--4                           & 0.85                 & 0                    & 0.91                 & 0.83                 & 0.57                 & 0.89                 & 0.59                 & 0.95                 & 0.99                    & 0.38                  \\
3--6                           & 0.88                 & 0                    & 0.90                  & 0.87                 & 0.61                 & 0.90                 & 0.59                 & 0.95                 & 0.99                    & 0.66                  \\
3--8                           & 0.87                 & 0                    & 0.91                 & 0.85                 & 0.60                  & 0.91                 & 0.59                 & 0.94                 & 0.99                    & 0.64                  \\
4--2                           & 0.83                 & 0                    & 0.89                 & 0.80                  & 0.55                 & 0.90                  & 0.97                 & 0.89                 & 0.99                    & 0                     \\
4--4                           & 0.87                 & 0                    & 0.90                  & 0.80                  & 0.60                  & 0.90                  & 0.98                 & 0.94                 &0.99                   & 0                     \\
4--6                           & 0.89                 & 0.19                 & 0.91                 & 0.83                 & 0.56                 & 0.90                  & 0.99                 & 0.94                 & 0.99                    & 0.21                  \\
4--8                           & 0.83                 & 0                    & 0.90                  & 0.78                 & 0                    & 0.87                 & 0.98                 & 0.95                 & 0.80                  & 0                     \\
5--2                           & 0                    & 0.36                 & 0.90                  & 0.74                 & 0.52                 & 0.88                 & 0.56                 & 0.83                 & 0.99                    & 0                     \\
5--4                           & 0.91                 & 0.38                 & 0.90                  & 0.76                 & 0.62                 & 0.90                  & 0.92                 & 0                    & 0.75                 & 0                     \\
5--6                           & 0.87                 & 0.36                 & 0.90                  & 0.79                 & 0.54                 & 0.87                 & 0.88                 & 0                    & 0.99                    & 0                     \\
5--8                           & 0.86                 & 0                    & 0.89                 & 0.80                 & 0                    & 0.86                 & 0.97                 & 0.85                 & 0.99                    & 0                     \\
6--2                           & 0.79                 & 0                    & 0.83                 & 0.71                 & 0                    & 0.82                 & 0.81                 & 0                    & 0.99                    & 0                     \\
6--4                           & 0.84                 & 0                    & 0.86                 & 0.73                 & 0                    & 0.88                 & 0.88                 & 0                    & 0.77                 & 0                     \\
6--6                           & 0                    & 0                    & 0.84                 & 0.59                 & 0                    & 0.86                 & 0.83                 & 0                    & 0.75                 & 0                     \\
6--8                           & 0                    & 0                    & 0.86                 & 0                    & 0                    & 0.85                 & 0.89                 & 0                    & 0.73                 & 0                     \\ 
\hline
\multicolumn{1}{c}{L-A} & clf                  & compound             & conj                 & cop                  & dep                  & det                  & discourse:sp         & flat                 & flat:name            & mark                  \\ 
\hline
2--2                           & 0.87                 & 0.86                 & 0.99                   & 0.88                 & 0.64                 & 0.97                 & 0.22                 & 0.96                 & 0.88                 &0.99                    \\
2--4                           & 0.82                 & 0.86                 & 0.99                 & 0.95                 & 0.63                 & 0.97                 & 0.22                 & 0.99                   & 0.88                 & 0.99                     \\
2--6                           & 0.89                 & 0.87                 & 0.99                 & 0.97                 & 0.66                 & 0.97                 & 0.29                 & 0.96                 & 0.88                 & 0.98                  \\
2--8                           & 0.83                 & 0.87                 & 0.99                 & 0.98                 & 0.62                 & 0.97                 & 0.33                 & 0.99                 & 0.88                 & 0.99                  \\
3--2                           & 0.88                 & 0.87                 & 0.99                    & 0.94                 & 0.64                 & 0.97                 & 0.22                 & 0.96                 & 0.92                 & 0.90                   \\
3--4                           & 0.86                 & 0.85                 & 0.99                    & 0.95                 & 0.64                 & 0.97                 & 0.20                  & 0.96                 & 0.94                 & 0.96                  \\
3--6                           & 0.88                 & 0.86                 & 0.99                   & 0.97                 & 0.66                 & 0.97                 & 0.21                    & 0.96                 & 0.94                 & 0.96                  \\
3--8                           & 0.90                  & 0.87                 & 0.99                 & 0.97                 & 0.66                 & 0.97                 & 0.22                 & 0.92                 & 0.97                 & 0.96                  \\
4--2                           & 0.68                 & 0.82                 & 0.97                 & 0.91                 & 0.64                 & 0.95                 & 0.18                 & 0.96                 & 0                    & 0.95                  \\
4--4                           & 0.66                 & 0.82                 & 0.99                    & 0.97                 & 0.65                 & 0.95                 & 0.22                 & 0.99                   & 0                    & 0.98                  \\
4--6                           & 0.69                 & 0.84                 & 0.99                 & 0.97                 & 0.68                 & 0.97                 & 0.29                 & 0.99                    & 0                    & 0.92                  \\
4--8                           & 0                    & 0.78                 & 0                    & 0.92                 & 0.64                 & 0.85                 & 0                    & 0.76                 & 0                    & 0.96                  \\
5--2                           & 0                    & 0.83                 & 0.99                    & 0.91                 & 0.64                 & 0.84                 & 0.33                 & 0.99                    & 0                    & 0.93                  \\
5--4                           & 0                    & 0.81                 & 0                    & 0.97                 & 0                    & 0.84                 & 0.29                 & 0.99                    & 0.80                  & 0.88                  \\
5--6                           & 0                    & 0.82                 & 0.99                   & 0.95                 & 0                    & 0.85                 & 0                    & 0.99                    & 0                    & 0.91                  \\
5--8                           & 0                    & 0.83                 & 0.86                 & 0.97                 & 0                    & 0.85                 & 0.22                 & 0.81                 & 0.84                 & 0.84                  \\
6--2                           & 0                    & 0.83                 & 0.53                 & 0.92                 & 0                    & 0.85                 & 0                    & 0.96                 & 0                    & 0.82                  \\
6--4                           & 0                    & 0.76                 & 0                    & 0.94                 & 0                    & 0.83                 & 0                    & 0.73                 & 0                    & 0.87                  \\
6--6                           & 0                    & 0.66                 & 0                    & 0.91                 & 0                    & 0.82                 & 0                    & 0.88                 & 0                    & 0.82                  \\
6--8                           & 0                    & 0.62                 & 0                    & 0.92                 & 0                    & 0.83                 & 0                    & 0.81                 & 0.72                 & 0.84                  \\ 
\hline
\multicolumn{1}{c}{L-A} & mark:prt             & mark:relcl           & nmod                 & nsubj                & nummod               & obj                  & obl                  & obl:tmod             & punct                & root                  \\ 
\hline
2--2                           & 0.68                 & 0.96                 & 0.92                 & 0.64                 & 0.97                 & 0.53                 & 0.79                 & 0.40                  & 0.99                    & 0.98                  \\
2--4                           & 0.66                 & 0.97                 & 0.93                 & 0.66                 & 0.98                 & 0.58                 & 0.79                 & 0.42                 &0.99                    & 0.98                  \\
2--6                           & 0.71                 & 0.97                 & 0.92                 & 0.68                 & 0.98                 & 0.61                 & 0.77                 & 0.44                 & 0.99                   & 0.98                  \\
2--8                           & 0.70                 & 0.97                 & 0.92                 & 0.67                 & 0.98                 & 0.59                 & 0.80                 & 0.41                 & 0.99                 & 0.98                  \\
3--2                           & 0.75                 & 0.98                 & 0.92                 & 0.68                 & 0.98                 & 0.63                 & 0.81                 & 0.42                 &0.99                    & 0.99                  \\
3--4                           & 0.73                 & 0.74                 & 0.73                 & 0.66                 & 0.99                 & 0.58                 & 0.84                 & 0.44                 & 0.99                   & 0.98                  \\
3--6                           & 0.69                 & 0.77                 & 0.72                 & 0.66                 & 0.99                 & 0.60                  & 0.79                 & 0.42                 & 0.99                  & 0.98                  \\
3--8                          & 0.69                 & 0.79                 & 0.71                 & 0.68                 & 0.99                 & 0.63                 & 0.84                 & 0.53                 & 0.99                    & 0.99                  \\
4--2                           & 0                    & 0.97                 & 0.92                 & 0.64                 & 0.97                 & 0.55                 & 0.80                  & 0.34                 & 0.99                   & 0.99                  \\
4--4                           & 0                    & 0.96                 & 0.94                 & 0.69                 & 0.99                 & 0.62                 & 0.82                 & 0.37                 & 0.99                   & 0.98                  \\
4--6                           & 0.72                 & 0.97                 & 0.92                 & 0.67                 & 0.99                 & 0.60                  & 0.82                 & 0.44                 & 0.99                   & 0.99                     \\
4--8                           & 0                    & 0.97                 & 0.90                  & 0.62                 & 0.98                 & 0.44                 & 0.78                 & 0.34                 & 0.98                 & 0.98                  \\
5--2                           & 0                    & 0.62                 & 0.72                 & 0.65                 & 0.98                 & 0.56                 & 0                    & 0.36                 & 0.99                 & 0.98                  \\
5--4                           & 0                    & 0.97                 & 0.92                 & 0.66                 & 0.86                 & 0.60                  & 0.77                 & 0                    & 0.99                    & 0.99                  \\
5--6                           & 0                    & 0.97                 & 0.91                 & 0.65                 & 0.85                 & 0.58                 & 0.73                 & 0.37                 & 0.99                 & 0.98                  \\
5--8                           & 0                    & 0.97                 & 0.92                 & 0.56                 & 0.83                 & 0.52                 & 0.73                 & 0                    & 0.99                   & 0.89                  \\
6--2                           & 0                    & 0.97                 & 0.89                 & 0.42                 & 0.83                 & 0                    & 0                    & 0                    & 0.98                 & 0                     \\
6--4                           & 0                    & 0.97                 & 0.90                  & 0.50                  & 0.86                 & 0                    & 0.64                 & 0                    & 0.98                 & 0.82                  \\
6--6                           & 0                    & 0.88                 & 0.68                 & 0.47                 & 0                    & 0                    & 0.66                 & 0                    & 0.96                 & 0.88                  \\
6--8                           & 0                    & 0.72                 & 0.80                  & 0.51                 & 0                    & 0                    & 0.66                 & 0                    & 0.99                 & 0.79                  \\ 
\hline
\multicolumn{1}{l}{}          & \multicolumn{1}{l}{} & \multicolumn{1}{l}{} & \multicolumn{1}{l}{} & \multicolumn{1}{l}{} & \multicolumn{1}{l}{} & \multicolumn{1}{l}{} & \multicolumn{1}{l}{} & \multicolumn{1}{l}{} & \multicolumn{1}{l}{} & \multicolumn{1}{l}{} 
\end{tabular}
\centering
\caption{GAT predictions of syntactic dependency in Chinese.}
\label{GATandBERT-1} 
\end{table*}

\begin{table*}[!ht]
\centering
\small
\begin{tabular}{c|c} 
\hline
\multicolumn{2}{c}{Zh}                 \\
\multicolumn{1}{c}{L-A} & xcomp  \\ 
\hline
2--2                           & 0.48   \\
2--4                           & 0.54   \\
2--6                           & 0.56   \\
2--8                           & 0.58   \\
3--2                           & 0.63   \\
3--4                           & 0.53   \\
3--6                           & 0.65   \\
3--8                           & 0.68   \\
4--2                           & 0.47   \\
4--4                           & 0.44   \\
4--6                           & 0.56   \\
4--8                           & 0.47   \\
5--2                           & 0.41   \\
5--4                           & 0.53   \\
5--6                           & 0.48   \\
5--8                           & 0      \\
6--2                           & 0      \\
6--4                           & 0      \\
6--6                           & 0      \\
6--8                           & 0      \\
\hline
\end{tabular}
\caption{GAT predictions of syntactic dependency in Chinese.}
\label{GATandBERT-2} 
\end{table*}

\begin{table*}[!ht]
\centering
\small
\begin{tabular}{c|cccccccccc} 
\hline
\multicolumn{11}{c}{Ru}                                                                                                          \\
\multicolumn{1}{c}{L-A} & acl   & acl:relcl & advcl  & advmod     & amod & appos & aux   & aux:pass    & case      & cc    \\ 
\hline
2--2                          & 0.54  & 0         & 0      & 0.90       & 0.98 & 0.32  & 0.75  & 0.96        & 0.99      & 0.97  \\
2--4                          & 0.52  & 0         & 0.71   & 0.91       & 0.98 & 0.55  & 0.89  & 0.96        & 0.99      & 0.99  \\
2--6                          & 0.64  & 0.81      & 0      & 0.89       & 0.98 & 0.24  & 0     & 0           & 0.98      & 0.96  \\
2--8                          & 0.64  & 0         & 0      & 0.90        & 0.98 & 0.50   & 0.67  & 0.92        & 0.98      & 0.97  \\
3--2                          & 0.57  & 0         & 0      & 0.90        & 0.98 & 0.12  & 0     & 0           & 0.98      & 0.96  \\
3--4                          & 0.63  & 0         & 0.56   & 0.92       & 0.98 & 0.45  & 0     & 0           & 0.98      & 0.96  \\
3--6                          & 0.63  & 0.84      & 0      & 0.90        & 0.98 & 0.48  & 0     & 0           & 0.98      & 0.96  \\
3--8                          & 0.67  & 0.72      & 0      & 0.91       & 0.98 & 0.13  & 0     & 0           & 0.99      & 0.96  \\
4--2                          & 0.51  & 0         & 0      & 0.92       & 0.97 & 0     & 0     & 0           & 0.97      & 0.84  \\
4--4                          & 0.60   & 0.64      & 0      & 0.89       & 0.97 & 0     & 0.67  & 0           & 0.99      & 0.82  \\
4--6                          & 0.73  & 0.84      & 0.39   & 0.90        & 0.98 & 0.65  & 0     & 0.86        & 0.99         & 0.82  \\
4--8                          & 0.65  & 0         & 0      & 0.92       & 0.99 & 0.55  & 0.44  & 0           & 0.99      & 0.96  \\
5--2                          & 0.57  & 0         & 0.23   & 0.91       & 0.96 & 0     & 0     & 0           & 0.97      & 0.85  \\
5--4                          & 0.67  & 0.78      & 0.49   & 0.91       & 0.97 & 0     & 0     & 0           & 0.98      & 0.82  \\
5--6                          & 0.77  & 0.75      & 0.17   & 0.91       & 0.97 & 0.44  & 0     & 0           & 0.97      & 0.81  \\
5--8                          & 0.56  & 0         & 0      & 0.91       & 0.96 & 0.54  & 0     & 0.86        & 0.99      & 0.86  \\
6--2                          & 0     & 0         & 0      & 0.90        & 0.96 & 0     & 0     & 0.89        & 0.94      & 0.83  \\
6--4                          & 0     & 0.42      & 0      & 0.88       & 0.88 & 0     & 0     & 0           & 0.95      & 0.78  \\
6--6                          & 0.30   & 0         & 0      & 0.88       & 0.91 & 0     & 0     & 0           & 0.94      & 0.79  \\
6--8                          & 0     & 0         & 0      & 0.90        & 0.96 & 0     & 0     & 0           & 0.96      & 0.85  \\ 
\hline
\multicolumn{1}{c}{L-A} & ccomp & conj      & cop    & csubj      & det  & fixed & flat  & flat:forign & flat:name & mark  \\ 
\hline
2--2                          & 0.70   & 0.84      & 0.96   & 0          & 0.99 & 0.43  & 0.85  & 0.87        & 0.58      & 0.97  \\
2--4                          & 0.67  & 0.87      & 0.99      & 0          & 0.99    & 0.57  & 0.86  & 0.92        & 0.56      & 0.94  \\
2--6                          & 0.54  & 0.88      & 0.58   & 0          & 0.98 & 0.61     & 0.87     & 0.80         & 0.52      & 0.96  \\
2--8                          & 0.57  & 0.87      & 0.96   & 0          & 0.99    & 0.50   & 0.86  & 0.87        & 0.64      & 0.90   \\
3--2                          & 0.50   & 0.88      & 0.56   & 0          & 0.98 & 0     & 0     & 0.74        & 0.51      & 0.93  \\
3--4                          & 0.81  & 0.90       & 0.67   & 0          & 0.99 & 0.67  & 0.86  & 0.87        & 0.55      & 0.94  \\
3--6                          & 0.67  & 0.89      & 0.67   & 0          & 0.99 & 0.56  & 0.77  & 0.83        & 0.59      & 0.93  \\
3--8                          & 0.63  & 0.87      & 0.65   & 0          & 0.99 & 0.67  & 0.86  & 0.92        & 0.61      & 0.93  \\
4--2                          & 0.60   & 0         & 0.63   & 0          & 0.99 & 0     & 0     & 0.69        & 0.52      & 0.94  \\
4--4                          & 0.31  & 0         & 0.73   & 0          & 0.99 & 0.76  & 0.77  & 0.83        & 0.64      & 0.94  \\
4--6                          & 0     & 0         & 0.96   & 0.13       & 0.99    & 0.84  & 0.67  & 0.83        & 0.69      & 0.97  \\
4--8                          & 0.72  & 0.88      & 0.85   & 0          & 0.99 & 0.80   & 0.80   & 0.92        & 0.68      & 0.94  \\
5--2                          & 0.63  & 0         & 0.56   & 0          & 0.99 & 0     & 0.55  & 0.88        & 0.59      & 0.93  \\
5--4                          & 0.69  & 0         & 0.58   & 0          & 0.99    & 0.71  & 0.77  & 0.87        & 0.59      & 0.96  \\
5--6                          & 0     & 0         & 0.61   & 0          & 0.99 & 0     & 0.67  & 0.80         & 0.62      & 0.93  \\
5--8                          & 0.49  & 0         & 0.96   & 0          & 0.99 & 0.80   & 0.48  & 0           & 0.61      & 0.96  \\
6--2                          & 0.28  & 0         & 0.88   & 0          & 0    & 0     & 0     & 0.71        & 0.58      & 0.91  \\
6--4                          & 0.48  & 0         & 0.63   & 0          & 0.94 & 0     & 0     & 0.81        & 0.43      & 0.97  \\
6--6                          & 0     & 0         & 0.58   & 0          & 0.93 & 0     & 0     & 0.74        & 0.43      & 0.93  \\
6--8                          & 0.49  & 0         & 0.56   & 0          & 0.99 & 0     & 0     & 0.83        & 0.55      & 0.93  \\ 
\hline
\multicolumn{1}{c}{L-A} & nmod  & nsubj     & nummod & nummod:gov & obj  & obl   & punct & root        & xcomp     &       \\ 
\hline
2--2                          & 0.90   & 0.71      & 0.76   & 0.33       & 0.58 & 0.89  & 0.99     & 0.98        & 0.53      &       \\
2--4                          & 0.90   & 0.67      & 0.75   & 0.43       & 0.56 & 0.91  & 0.99     & 0.98        & 0.53      &       \\
2--6                          & 0.88  & 0.67      & 0.76   & 0          & 0.48 & 0.90   & 0.99     & 0.98        & 0         &       \\
2--8                          & 0.90   & 0.69      & 0.75   & 0          & 0.54 & 0.91  & 0.99     & 0.98        & 0         &       \\
3--2                          & 0.88  & 0.67      & 0.65   & 0.31       & 0.55 & 0.93  & 0.99     & 0.98        & 0         &       \\
3--4                          & 0.89  & 0.69      & 0.71   & 0.43       & 0.59 & 0.92  & 0.99     & 0.99        & 0.56      &       \\
3--6                          & 0.91  & 0.67      & 0.73   & 0.50        & 0.52 & 0.92  & 0.99     & 0.98        & 0         &       \\
3--8                          & 0.91  & 0.70       & 0.71   & 0.40        & 0.60  & 0.93  & 0.99     & 0.99        & 0         &       \\
4--2                          & 0.83  & 0.70       & 0.70    & 0.43       & 0.57 & 0.90   & 0.99  & 0.94        & 0.45      &       \\
4--4                          & 0.86  & 0.65      & 0.71   & 0.43       & 0.52 & 0.91  & 0.99  & 0           & 0         &       \\
4--6                          & 0.91  & 0.72      & 0.75   & 0.43       & 0.59 & 0.92  &0.99     & 0.98        & 0         &       \\
4--8                          & 0.92  & 0.71      & 0.77   & 0.40        & 0.63 & 0.93  & 0.99     & 0.98        & 0.61      &       \\
5--2                          & 0.87  & 0.63      & 0.78   & 0.53       & 0.44 & 0.90   & 0.99  & 0           & 0         &       \\
5--4                          & 0.83  & 0.71      & 0.72   & 0.31       & 0.56 & 0.90   & 0.99  & 0.97        & 0.52      &       \\
5--6                          & 0.87  & 0.69      & 0.72   & 0.31       & 0.60  & 0.89  & 0.99  & 0           & 0.52      &       \\
5--8                          & 0.89  & 0.68      & 0.79   & 0.43       & 0.50  & 0.91  & 0.99  & 0.98        & 0         &       \\
6--2                          & 0.78  & 0.67      & 0.68   & 0          & 0.41 & 0.88  & 0.98  & 0.96        & 0         &       \\
6--4                          & 0     & 0.64      & 0.62   & 0          & 0.46 & 0.75  & 0.99  & 0.95        & 0         &       \\
6--6                          & 0     & 0.53      & 0.54   & 0          & 0.40  & 0.75  & 0.98  & 0           & 0         &       \\
6--8                          & 0.83  & 0.53      & 0.63   & 0          & 0.40  & 0.88  & 0.99  & 0           & 0         &       \\
\hline
\end{tabular}
\caption{GAT predictions of syntactic dependency in Russian.}
\label{GATandBERT-3} 
\end{table*}

\begin{table*}[!ht]
\centering
\small
\begin{tabular}{c|ccccccccc} 
\hline
\multicolumn{10}{c}{De}                                                                                                  \\
\multicolumn{1}{c}{L-A} & acl  & acl:relel & advcl    & advmod       & amod & appos & aux      & aux:pass  & case  \\ 
\hline
2--2                          & 0    & 0.71      & 0.83     & 0.99         & 0.95 & 0.39  & 0.85     & 0.81      & 0.99  \\
2--4                          & 0.5  & 0.75      & 0.89     & 0.99         & 0.95 & 0.56  & 0.91     & 0.81      & 0.99  \\
2--6                          & 0.5  & 0.75      & 0.89     & 0.99         & 0.95 & 0.56  & 0.91     & 0.81      & 0.99  \\
2--8                          & 0    & 0.41      & 0        & 0.99         & 0.94 & 0     & 0.86     & 0.81      & 0.99  \\
3--2                          & 0    & 0.60       & 0        & 0.99         & 0.94 & 0     & 0.85     & 0.81      & 0.99  \\
3--4                          & 0    & 0.45      & 0        & 0.99         & 0.94 & 0     & 0.85     & 0.81      & 0.99  \\
3--6                          & 0    & 0.41      & 0        & 0.98         & 0.94 & 0     & 0.88     & 0.81      & 0.99  \\
3--8                          & 0    & 0.46      & 0        & 0.99         & 0.94 & 0     & 0.88     & 0.81      & 0.99  \\
4--2                          & 0    & 0.52      & 0        & 0.99            & 0.95 & 0     & 0.81     & 0         & 0.99  \\
4--4                          & 0    & 0.45      & 0        & 0.99         & 0.94 & 0     & 0        & 0         & 0.99  \\
4--6                          & 0    & 0.40       & 0        & 0.98         & 0.93 & 0     & 0        & 0.48      & 0.99  \\
4--8                          & 0    & 0.45      & 0        & 0.98         & 0.93 & 0     & 0        & 0.52      & 0.99  \\
5--2                          & 0    & 0.42      & 0        & 0.99         & 0.92 & 0     & 0.86     & 0.81      & 0.99  \\
5--4                          & 0    & 0.68      & 0        & 0.99         & 0.93 & 0     & 0.85     & 0.81      & 0.99  \\
5--6                          & 0    & 0.44      & 0        & 0.99         & 0.94 & 0     & 0        & 0         & 0.99  \\
5--8                          & 0    & 0.43      & 0        & 0.97         & 0.94 & 0     & 0        & 0         & 0.99  \\
6--2                          & 0    & 0         & 0        & 0.98         & 0.90  & 0.07  & 0.62     & 0         & 0.98  \\
6--4                          & 0    & 0         & 0        & 0.97         & 0.91 & 0     & 0        & 0.70       & 0.98  \\
6--6                          & 0    & 0         & 0        & 0.97         & 0.91 & 0     & 0        & 0         & 0.98  \\
6--8                          & 0    & 0.37      & 0        & 0.97         & 0.91 & 0     & 0        & 0         & 0.98  \\ 
\hline
\multicolumn{1}{c}{L-A} & cc   & ccomp     & compound & compound:prt & conj & cop   & det      & flat:name & mark  \\ 
\hline
2--2                          & 0.99    & 0.56      & 0.80      & 0            & 0.78 & 0.93  & 0.99     & 0.83      & 0.97  \\
2--4                          & 0.99    & 0.60       & 0.81     & 0            & 0.81 & 0.98  & 0.99     & 0.85      & 0.97  \\
2--6                          &0.99    & 0.60       & 0.81     & 0            & 0.81 & 0.98  & 0.99     & 0.85      & 0.97  \\
2--8                          &0.99    & 0         & 0.72     & 0            & 0.82  & 0.95  & 0.99     & 0.81      & 0.96  \\
3--2                          & 0.99 & 0.48      & 0.83     & 0            & 0.78 & 0.93  & 0.99     & 0.82      & 0.95  \\
3--4                          & 0.99 & 0         & 0.80      & 0            & 0.80  & 0.95  & 0.99     & 0.84      & 0.86  \\
3--6                          & 0.99 & 0         & 0.78     & 0            & 0.80  & 0.95  & 0.99     & 0.81      & 0.91  \\
3--8                          & 0.99 & 0         & 0.72     & 0            & 0.80  & 0.95  & 0.99     & 0.84      & 0.91  \\
4--2                          & 0.99   & 0         & 0.86     & 0            & 0.76 & 0.93  & 0.99     & 0.90       & 0.93  \\
4--4                          & 0.99 & 0         & 0.82     & 0            & 0.79 & 0.57  & 0.99     & 0.82      & 0.84  \\
4--6                          & 0.99 & 0         & 0.76     & 0            & 0.79 & 0.90   & 0.99     & 0.85      & 0.93  \\
4--8                          & 0.99 & 0         & 0.80      & 0            & 0.80  & 0.88  & 0.99     & 0.84      & 0.85  \\
5--2                          & 0.99 & 0         & 0.82     & 0            & 0.82 & 0.95  & 0.99     & 0.83      & 0.92  \\
5--4                          & 0.99 & 0.52      & 0.74     & 0            & 0.82 & 0.95  & 0.99     & 0.8       & 0.94  \\
5--6                          & 0.99 & 0         & 0.75     & 0            & 0.82 & 0.65  & 0.99     & 0.78      & 0.85  \\
5--8                          & 0.99 & 0         & 0        & 0            & 0.79 & 0.57  & 0.99     & 0.78      & 0.86  \\
6--2                          & 0.98 & 0         & 0.65     & 0.67         & 0.74 & 0     & 0.96     & 0.84      & 0.82  \\
6--4                          & 0.99 & 0         & 0.69     & 0            & 0.78 & 0.70   & 0.97     & 0.83      & 0.84  \\
6--6                          & 0.99 & 0         & 0.63     & 0.69         & 0.68 & 0.54  & 0.98     & 0.71      & 0.81  \\
6--8                          & 0.93 & 0         & 0.71     & 0            & 0    & 0.55  & 0.99     & 0.73      & 0.87  \\ 
\hline
\multicolumn{1}{c}{L-A} & nmod & nmod:poss & nsubj    & nummod       & obj  & obl   & obl:tmod & punct     & root  \\ 
\hline
2--2                          & 0.82 & 0.85      & 0.75     & 0.84         & 0.63 & 0.80   & 0        & 0.99         & 0.96  \\
2--4                          & 0.83 & 0.88      & 0.72     & 0.84         & 0.63 & 0.83  & 0        & 0.99         & 0.97  \\
2--6                          & 0.83 & 0.88      & 0.72     & 0.84         & 0.63 & 0.83  & 0        & 0.99        & 0.97  \\
2--8                          & 0.76 & 0.86      & 0.69     & 0.84         & 0.56 & 0.80  & 0        & 0.99         & 0.94  \\
3--2                          & 0.80  & 0.85      & 0.78     & 0.87         & 0.67 & 0.84  & 0        & 0.99         & 0.97  \\
3--4                          & 0.80  & 0.86      & 0.71     & 0.84         & 0.37 & 0.84  & 0        &0.99        & 0.92  \\
3--6                          & 0.79 & 0.85      & 0.72     & 0.87         & 0.56 & 0.86  & 0        & 0.99         & 0.93  \\
3--8                          & 0.81 & 0.83      & 0.74     & 0.87         & 0.59 & 0.84  & 0        & 0.99         & 0.93  \\
4--2                          & 0.81 & 0.86      & 0.74     & 0.84         & 0.65 & 0.85  & 0        & 0.99      & 0.95  \\
4--4                          & 0.78 & 0.85      & 0.73     & 0.87         & 0.51 & 0.86  & 0        &0.99        & 0.93  \\
4--6                          & 0.81 & 0.82      & 0.77     & 0.84         & 0.65 & 0.85  & 0        & 0.99         & 0.93  \\
4--8                          & 0.78 & 0.86      & 0.74     & 0.87         & 0.64 & 0.86  & 0        & 0.99         & 0.95  \\
5--2                          & 0.81 & 0.83      & 0.78     & 0.90          & 0.62 & 0.83  & 0.44     & 0.99      & 0.89  \\
5--4                          & 0.82 & 0.84      & 0.79     & 0.90          & 0.66 & 0.87  & 0.44     & 0.99      & 0.96  \\
5--6                          & 0.82 & 0.85      & 0.72     & 0.87         & 0.56 & 0.82  & 0        & 0.99      & 0.96  \\
5--8                          & 0.76 & 0.83      & 0.73     & 0.80          & 0.60  & 0.85  & 0        & 0.97      & 0.89  \\
6--2                          & 0.73 & 0.81      & 0.65     & 0.67         & 0.23 & 0.72  & 0        & 0.97      & 0.89  \\
6--4                          & 0.75 & 0.85      & 0.65     & 0.76         & 0.23 & 0.87  & 0        & 0.97      & 0.79  \\
6--6                          & 0.81 & 0.85      & 0.67     & 0.81         & 0.22 & 0.85  & 0        & 0.98      & 0.90   \\
6--8                          & 0.66 & 0         & 0.63     & 0.81         & 0    & 0.86  & 0        & 0.98      & 0.89  \\
\hline
\end{tabular}
\caption{GAT predictions of syntactic dependency in German.}
\label{GATandBERT-4} 
\end{table*}

\begin{table*}[!ht]
\centering
\small
\begin{tabular}{c|c} 
\hline
\multicolumn{2}{c}{De}                 \\
\multicolumn{1}{c}{L-A} & xcomp  \\ 
\hline
2--2                          & 0.55   \\
2--4                          & 0.49   \\
2--6                          & 0.49   \\
2--8                          & 0      \\
3--2                          & 0.38   \\
3--4                          & 0      \\
3--6                          & 0      \\
3--8                          & 0      \\
4--2                          & 0.41   \\
4--4                          & 0      \\
4--6                          & 0      \\
4--8                          & 0      \\
5--2                          & 0      \\
5--4                          & 0      \\
5--6                          & 0      \\
5--8                          & 0      \\
6--2                          & 0      \\
6--4                          & 0      \\
6--6                          & 0      \\
6--8                          & 0      \\
\hline
\end{tabular}
\caption{GAT predictions of syntactic dependency in German.}
\label{GATandBERT-5} 
\end{table*}
\end{document}